\documentclass{article}

\usepackage{microtype}
\usepackage{graphicx}
\usepackage{subcaption}
\usepackage{booktabs} 

\usepackage{hyperref}

 \usepackage[preprint]{icml2026}

\usepackage{amsmath}
\usepackage{amssymb}
\usepackage{mathtools}
\usepackage{amsthm}
\usepackage{multirow}

\usepackage[capitalize,noabbrev]{cleveref}

\theoremstyle{plain}

\theoremstyle{definition}

\theoremstyle{remark}

\usepackage[textsize=tiny]{todonotes}

\icmltitlerunning{GVP-WM: Grounding Generated Videos in Feasible Plans via World Models}
\setcitestyle{numbers,square,comma,sort&compress}

\begin{document}

\twocolumn[
  \icmltitle{Grounding Generated Videos in Feasible Plans via World Models}



  \icmlsetsymbol{equal}{*}

  \begin{icmlauthorlist}
    \icmlauthor{Christos Ziakas}{sch}
    \icmlauthor{Amir Bar}{comp}
    \icmlauthor{Alessandra Russo}{sch}
  \end{icmlauthorlist}

  \icmlaffiliation{comp}{UC Berkeley}
  \icmlaffiliation{sch}{Imperial College London}

  \icmlcorrespondingauthor{Christos Ziakas}{c.ziakas24@imperial.ac.uk}

  \icmlkeywords{Machine Learning, ICML}

  \vskip 0.3in
]



\printAffiliationsAndNotice{}  

\newcommand{\amir}[1]{{\textcolor{red}{amir: #1}}}
\newcommand{\ale}[1]{{\textcolor{blue}{ale: #1}}}
\newcommand{\christos}[1]{{\textcolor{orange}{christos: #1}}}

\begin{abstract}

Large-scale video generative models have shown emerging capabilities as zero-shot visual planners, yet video-generated plans often violate temporal consistency and physical constraints, leading to failures when mapped to executable actions. To this end, we propose Grounding Video Plans with World Models (GVP-WM), a planning method that grounds video-generated plans into feasible action sequences using a pre-trained action-conditioned world model. At test-time, GVP-WM first generates a video plan from initial and goal observations, then projects the video guidance onto the manifold of dynamically feasible latent trajectories via video-guided latent collocation. In particular, we formulate grounding as a goal-conditioned latent-space trajectory optimization problem that jointly optimizes latent states and actions under world-model dynamics, while preserving semantic alignment with the video-generated plan. Empirically, GVP-WM recovers feasible long-horizon plans from zero-shot image-to-video–generated and motion-blurred videos that violate physical constraints, across navigation and manipulation simulation tasks. Project website: \url{https://chziakas.github.io/gvpwm/}

\end{abstract}

\section{Introduction}
\label{sec:intro}

Large video generative models have demonstrated strong zero-shot capabilities in synthesizing realistic and temporally coherent videos across a wide range of domains~\citep{ho2022video, latentdiff,sora, wan2025}. Beyond video synthesis, diffusion video models can generate expert-like video plans, leveraging their capabilities in prompt instruction following, object permanence, and physical consistency \citep{unipi,ko2023learning,largevideoplans,dreamgen}. Recent work further suggests that diffusion-based video models may exhibit emergent visual reasoning capabilities~\citep{Geirhos}. However, generated videos are often physically infeasible (e.g., object teleportation) or temporally inconsistent (e.g., motion blur), leading to violations of real-world dynamics, particularly under out-of-distribution conditions~\citep{huang2024vbench,bansal2025videophy,nvidia2025cosmos}. Despite advances in diffusion-based video models for temporal coherence \citep{chen2024diffusionforcing,song2025history} and kinematics conditioning \citep{hu2023animate,bai2025peva}, inferring actions directly from video-generated plans may violate real-world dynamics \citep{largevideoplans, ni2025gemmimic}. Prior work leveraged video-generated plans as subgoals for model-predictive control \citep{nair2020hierarchical} and hierarchical reinforcement learning \citep{black2023zero}. However, these approaches assume that visual subgoals are feasible during execution, without accounting for the quality of the underlying video plans. In contrast, recent work~\citep{luo2025grounding} grounds video-generated plans via goal-conditioned exploration during policy learning, enabling divergence from video guidance to ensure feasibility at the cost of additional training with environment interaction.




\begin{figure*}[t]
\vskip 0.1in
\begin{center}
\centerline{\includegraphics[width=1.0\textwidth]{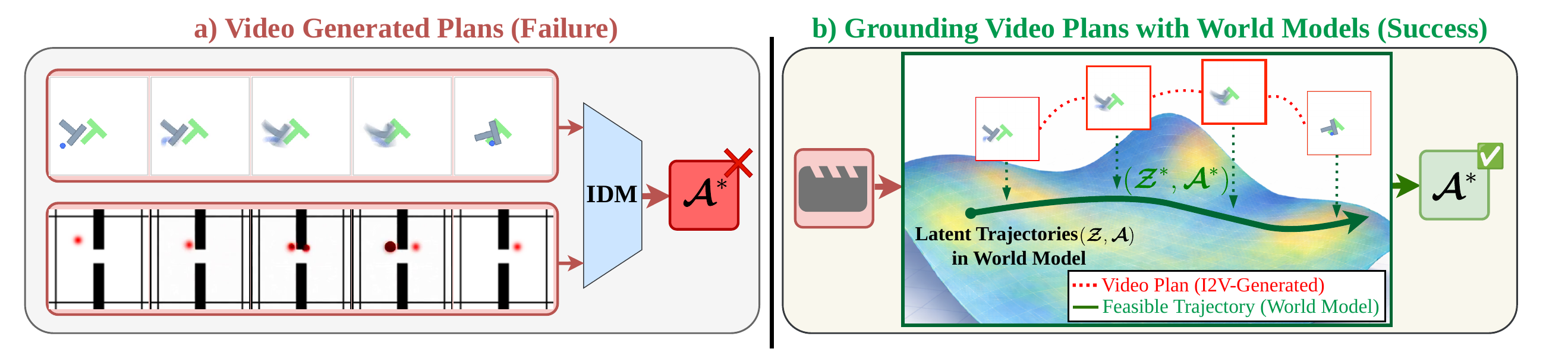}}

\caption{
\textbf{(a)} Zero-shot I2V-generated plans  violate temporal consistency and physical constraints, leading to failures when mapped to executable actions via inverse-dynamics models (IDM).
\textbf{(b)} GVP-WM projects video-generated plans onto dynamically feasible latent trajectories via latent-space trajectory optimization under a pre-trained action-conditioned world model.}

\label{fig:architecture}
\end{center}
\vskip -0.2in
\end{figure*}

In this work, we introduce Grounding Video Plans with World Models (GVP-WM), a planning method that grounds video-generated plans into feasible actions at test time using pre-trained action-conditioned world models~\citep{ha2018world}. GVP-WM projects video-generated plans onto the manifold of dynamically feasible latent trajectories under the dynamics of world models via video-guided latent collocation~\citep{latco}. Latent collocation formulates goal-conditioned planning as a trajectory optimization problem, optimizing a latent trajectory using video plans as guidance, subject to world-model dynamics constraints. In fact, GVP-WM jointly optimizes both latent states and actions, as latent states inferred from video plans may violate world-model dynamics constraints. GVP-WM incorporates video guidance during latent collocation by initializing the latent trajectory from the video plan and penalizing scale-invariant semantic deviation between the optimized latent states and the video plan. 

Our main contributions are as follows:

\begin{itemize}
 \item  We propose GVP-WM, a test-time method for grounding video-generated plans into physically feasible action sequences using a pre-trained action-conditioned world model.
 \item We formulate the grounding of video plans as a latent-space trajectory optimization problem under world-model dynamics, projecting video-generated plans onto feasible latent trajectories while preserving semantic alignment with the video plan.
 \item We empirically demonstrate that GVP-WM outperforms video-to-action inverse-dynamics models on navigation and manipulation simulation tasks under image-to-video–generated and motion-blurred video plans, particularly in long-horizon settings.

\end{itemize}

\section{Preliminaries}
\label{sec:formulation}


\subsection{Goal-Conditioned Planning}
\label{sec:goal_planning}

We formulate planning as a finite-horizon goal-conditioned Markov Decision Process~\citep{sutton1998reinforcement, finn2017deep}, defined by the tuple $\mathcal{M} = \langle \mathcal{S}, \mathcal{A}, p, \mathcal{G}, \mathcal{C}, \mathcal{T} \rangle$. 
$\mathcal{S}$ denotes the state space, $\mathcal{A}$ the continuous action space, and $\mathcal{G} \subset \mathcal{S}$ the goal space. The environment dynamics are defined by an unknown stochastic transition function $p(s_{t+1} \mid s_t, a_t)$. The goal-conditioned cost function $\mathcal{C}: \mathcal{S} \times \mathcal{A} \times \mathcal{G} \to \mathbb{R}$ assigns a scalar cost for taking action $a$ in state $s$ under goal state $g$. Given an initial state $s_0$ and a goal $g \in \mathcal{G}$, goal-conditioned planning finds an action sequence $a_{0:T-1}$ that minimizes $\mathbb{E} \left[ \sum_{t=0}^{T-1} C(s_t, a_t, g) \right]$ over a finite horizon $T$.



\subsection{Latent Collocation in World Models}
\label{sec:latent_planning}

Model-based planning approximates  environment dynamics using an action-conditioned world model in latent space, as planning directly in pixel space is computationally infeasible~\citep{dreamer,lecun2022path}. In particular, an action-conditioned latent world model consists of an encoder $E_\phi: \mathcal{S} \to \mathcal{Z}$ that maps observations to compact latent states $z_t = E_\phi(s_t)$, and a transition function $f_\psi$ that predicts future latent states conditioned on a history of $H$ latent states and actions $z_{t+1} = f_\psi(z_{t-H:t}, a_{t-H:t})$. Planning in latent world models minimizes a latent cost objective $\tilde{C}: \mathcal{Z} \times \mathcal{A} \times \mathcal{Z} \to \mathbb{R}$, defined over latent trajectories, measuring progress from the initial state $s_0$  to the goal state $g$ in latent space using $E_\phi$. In contrast to shooting-based or gradient-based planning methods, where intermediate latent states are implicitly determined by forward simulation of actions under the world model, latent collocation treats both latent states and actions as explicit decision variables~\citep{latco}. In particular, latent collocation formulates planning as a latent-space trajectory optimization problem that jointly optimizes latent (knot) states $z_{0:T}$ and actions $a_{0:T-1}$ to minimize $\tilde{C}$ subject to learned world-model dynamics constraints.

\section{Grounding Video Plans with World Models}
\label{sec:method}

\begin{figure*}[t]
\begin{center}
\centerline{\includegraphics[width=0.75\textwidth]{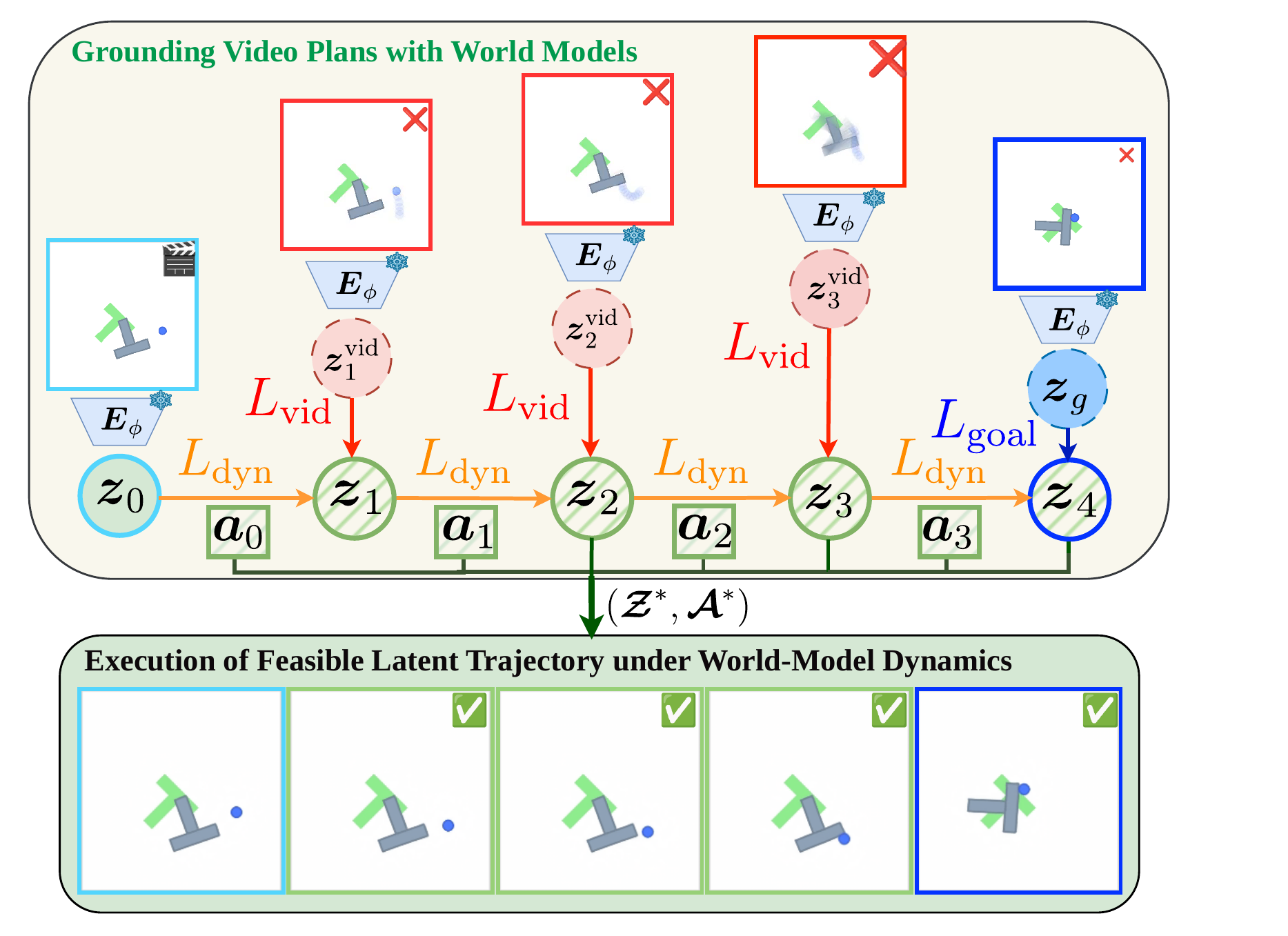}}

\caption{Overview of GVP-WM. A video plan, which may contain physically infeasible transitions (e.g., motion blur or object teleportation), is encoded into a sequence of latent states ${z^{\mathrm{vid}}_{t:T-1}}$ using a pretrained visual encoder $E\phi$ of the world model. Video-guided latent collocation optimizes a latent trajectory ${z_{t+1:T}}$ and corresponding actions ${a_{t:T-1}}$ by minimizing an augmented Lagrangian objective (Eq.~\ref{eq:auglag}), which balances video alignment ($L_{\mathrm{vid}}$), terminal goal reaching ($L_{\mathrm{goal}}$), and world-model dynamics ($L_{\mathrm{dyn}}$). The actions from the optimal latent trajectory satisfying the world-model dynamics constraints are executed using model predictive control.}

\label{fig:architecture}
\end{center}
\vskip -0.2in
\end{figure*}

Our method first generates a video plan from the initial and goal observations and then grounds it into a feasible action sequence using trajectory optimization in latent space. We provide the pseudocode for GVP-WM in Algorithm~\ref{alg:gvpwm_mpc}.

\subsection{Problem Formulation}
\label{sec:problem}


Video models may produce video plans with temporally inconsistent or blurry transitions, particularly in out-of-distribution settings~\citep{huang2024vbench,bansal2025videophy}. We consider the problem of grounding a video plan $\tau_{\mathrm{vid}}$ into a feasible trajectory of latent states $z_{0:T}$ and actions $a_{0:T-1}$ using an action-conditioned world model $(E_\phi, f_\psi)$. We formulate this task as a finite-horizon goal-conditioned MDP, defined in Section \ref{sec:goal_planning}, where the state space $\mathcal{S}$ consists of high-dimensional visual observations $o_t$ and, optionally, low-dimensional proprioceptive information $p_t$. The goal space $\mathcal{G}$ is defined over visual observations, where each goal $g \in \mathcal{G}$ corresponds to a target image $o_g$. Given an initial state $s_0$, a goal observation $o_g$, and a video-generated plan $\tau_{\mathrm{vid}}$, our objective is to optimize a latent trajectory that satisfies the world model dynamics $f_\psi$, while following the semantic guidance provided by the video plan $\tau_{\mathrm{vid}}$. In this paper, we formulate this grounding process as a constrained trajectory optimization problem in latent space, using video-guided latent collocation as described below.



\subsection{Video-Guided Latent Collocation}
\label{sec:grounding}

Our goal is to find a latent trajectory of $z_{0:T}$ and $a_{0:T-1}$ that is guided by a video plan $ \tau_{\text{vid}}$ while satisfying the environment transition dynamics of the world model $f_\psi$.

\subsubsection{Video Plan Generation}

A video plan $\tau_{\mathrm{vid}}$ is a sequence of visual observations generated by a conditional video generative model $\mathcal{G}$ that provides temporally ordered visual foresight for completing a task. $\mathcal{G}$ can be instantiated using an image-to-video (I2V) diffusion-based video model that produces temporally coherent visual transitions between the start and goal observations~\citep{blattmann2023stable}. In particular, I2V diffusion models with first--last frame conditioning employ a masking mechanism that fixes the initial and final frames as the start and goal observations, guiding the generation of intermediate frames while maintaining temporal consistency~\citep{wan2025}. Given an initial visual observation $o_0$ and a visual goal observation $o_g$, we use a diffusion-based I2V generation model $\mathcal{G}$ trained with first--last frame transformation to synthesize a video plan $ \tau_{\text{vid}}$ as 
\begin{equation}
    \tau_{\text{vid}} = \{o_0, \tilde{o}_1, \ldots, \tilde{o}_{T-1}, o_g\}
    \sim \mathcal{G}(\cdot \mid o_0, o_g, c),
\end{equation}
where $c$ denotes optional contextual information, such as a textual description of the task.
The video generation model can optionally be fine-tuned on a dataset of domain-specific demonstrations $\mathcal{D}_{\text{demo}}$ to improve physical consistency.  If the video generation model and the world model operate at different frame-skip ratios, temporal subsampling is applied to the video plan to ensure temporal alignment.

\subsubsection{Video Guidance}
\label{sec:video_guidance}

We map the generated video plan into the latent state space of the action-conditioned world model using its underlying visual encoder, yielding a sequence of latent states $z^{\text{vid}}_{0:t}$. However, the latent representations derived from generated video frames may exhibit magnitude drift relative to the in-distribution latent states of the pre-trained world model. To improve robustness under distribution shift, we opt for world models with pre-trained frozen visual encoders \citep{dino-wm}. In addition, we employ a semantic alignment loss between the optimized latent trajectory and the video plan, which is equivalent  to cosine similarity between their latent states.  Specifically, the video alignment loss is defined as
\begin{equation}
\label{eq:cosine_equiv}
\mathcal{L}_{\text{vid}}(z_t, z^{\text{vid}}_t)
= \left| \phi(z_t) - \phi(z^{\text{vid}}_t) \right|^2,
\end{equation}
where both optimized and video latent states are projected onto the unit $\ell_2$ hypersphere by $\phi(z) = z / \|z\|_2$. This loss penalizes angular deviation between latent embeddings while remaining invariant to their magnitude~\citep{grill2020bootstrap}.

\subsubsection{Grounding via Latent Collocation}

We formulate grounding video plans into feasible action sequences as a video-guided direct collocation problem. In direct collocation, both the latent states $\mathcal{Z} = {z_0, \ldots, z_T}$ and the actions $\mathcal{A} = {a_0, \ldots, a_{T-1}}$ are treated as decision variables, as described in Section~\ref{sec:latent_planning}. Our goal is to solve for a trajectory $(\mathcal{Z}^\ast, \mathcal{A}^\ast)$ that minimizes the divergence between the latent states of the optimized trajectory and the video plan $z^{\text{vid}}$, while satisfying the world model dynamics $f\psi$. To this end, we instantiate the latent cost objective $\tilde{C}$ defined in \ref{sec:latent_planning} as a weighted combination of a video alignment loss, a terminal goal loss, and an action regularization term, leading to the following constrained optimization problem: 
\begin{equation}
\label{eq:cons_prob}
\begin{aligned}
\min_{\mathcal{Z}, \mathcal{A}} \quad & 
\lambda_{\text{v}} \sum_{t=1}^{T-1} \mathcal{L}_{\text{vid}}(z_t, z^{\text{vid}}_t) + \lambda_{\text{g}} \mathcal{L}_{\text{goal}}(z_T, z_g) 
+ \lambda_{\text{r}} \sum_{t=0}^{T-1} \| a_t \|^2 \\
\text{s.t.} \quad & 
z_{t+1} = f_\psi(z_{t-H:t}, a_{t-H:t}), \quad \forall t \in \{0, \ldots, T-1\}, \\
& a_{\min} \preceq a_t \preceq a_{\max}, \quad \forall t \in \{0, \ldots, T-1\}.
\end{aligned}
\end{equation}
In our formulation, the dynamics of the latent action-conditioned world model $f_\psi$ are enforced as hard constraints. The video alignment loss $\mathcal{L}_{\text{vid}}$ corresponds to the objective defined in Eq.~\ref{eq:cosine_equiv}. The goal loss $\mathcal{L}_{\text{goal}}$ is defined as the mean squared error between the terminal latent state $z_T$ and the latent representation of a visual goal $z_g$. The coefficients $\lambda_v$, $\lambda_g$, and $\lambda_r$ are hyperparameters that balance the contribution of the loss terms. We empirically validate the impact of video guidance, latent collocation, and scale-invariant semantic alignment via ablation studies in Appendix~\ref{app:abl}.

\begin{algorithm}[t]
\caption{Grounding Video Plans with World Models}
\label{alg:gvpwm_mpc}
\begin{algorithmic}[1]
\REQUIRE initial state $s_0 = (o_0, p_0)$; goal observation $o_g$; video model $\mathcal{G}$; world model $(E_\phi, f_\psi)$;  MPC parameters $(T, K)$; ALM iteration and penalty parameters
\STATE Generate a video plan:
$\tau_{\mathrm{vid}} \sim \mathcal{G}(\cdot \mid o_0, o_g, c)$
\STATE Encode video plan into latent space:
$z^{\mathrm{vid}}_{0:T} \leftarrow E_\phi(\tau_{\mathrm{vid}})$
\STATE Initialize primal variables:
$\mathcal{Z} \leftarrow z^{\mathrm{vid}}_{0:T}$,\ \ $\mathcal{A} \leftarrow \mathbf{0}$
\FOR{$t = 0$ \textbf{to} $T-1$ \textbf{step} $K$}
    \STATE Set initial latent state of trajectory: $z_t \leftarrow E_\phi(s_t)$
    \STATE Initialize dual variables: $\lambda \leftarrow \mathbf{0}$, $\rho \leftarrow \rho_0$
    \FOR{$k = 1$ \textbf{to} $O_{\mathrm{ALM}}$}
        \FOR{$i = 1$ \textbf{to} $I_{\mathrm{ALM}}$} 
            \STATE Video Guidance:
            $\mathcal{L}_{\mathrm{vid}}(z_{t+1:T-1}, z^{\mathrm{vid}}_{t+1:T-1})$
            \STATE Terminal Latent Goal: $\mathcal{L}_{\mathrm{goal}}(z_T, z_g)$
            \STATE Dynamic Constraints:$\mathcal{L}_{\mathrm{dyn}}^{t}(z_{t-H:t+1}, a_{t-H:t}; f_\psi)$
            \STATE Compute Augmented Lagrangian $\mathcal{L}_{\rho}$ (Eq.\ref{eq:auglag})
            \STATE Primal update: $(\mathcal{Z}, \mathcal{A})$ via a gradient step on $\mathcal{L}_{\rho}$
        \ENDFOR
        \STATE Dual update:
        $\lambda \leftarrow \lambda + \rho \, \mathcal{L}_{\mathrm{dyn}}$
        \STATE Penalty update:
        $\rho \leftarrow \min(\gamma \rho, \rho_{\max})$
    \ENDFOR
    \STATE Execute $a_{t:t+K}$ from $\mathcal{A}$ with sampling refinement
    \STATE Update current state: $s_{t+K} \leftarrow  (o_{t+K}, p_{t+K})$
\ENDFOR
\end{algorithmic}
\end{algorithm}

\subsection{Visual Model Predictive Control}
At each timestep $t$, we solve the video-guided latent collocation optimization problem over the planning horizon, producing the optimal trajectory $(z^*_{t+1:T}, a^*_{t:T-1})$. We then execute actions with Model Predictive Control (MPC).

\subsubsection{Optimization via Augmented Lagrangian}
\label{sec:alm}

To efficiently solve the non-linear constrained optimization problem described in Equation \ref{eq:cons_prob}, we employ the Augmented Lagrangian Method (ALM)~\citep{bertsikas}. The loss is defined as:
\begin{equation}
\label{eq:auglag}
\mathcal{L}_\rho(\mathcal{Z}, \mathcal{A}, \Lambda) =\tilde{C}(\mathcal{Z}, \mathcal{A}) + \sum_{t=0}^{T-1} \left(\lambda_t^\top \mathcal{L}_{\mathrm{dyn}}^{t} + \frac{\rho}{2} \| \mathcal{L}_{\mathrm{dyn}}^{t} \|^2 \right),
\end{equation}
where $\tilde{C}$ denotes the latent cost objective from Eq.~\ref{eq:cons_prob}. $\Lambda = \{\lambda_0, \ldots, \lambda_{T-1}\}$ are the Lagrange multipliers, and $\rho > 0$ is a scalar penalty parameter. The term $\mathcal{L}_{\mathrm{dyn}}^{t}$ corresponds to the dynamics constraint violation at timestep $t$, defined as:
\begin{equation}
\label{eq:dynamic_loss}
\mathcal{L}_{\mathrm{dyn}}^{t}(\mathcal{Z}, \mathcal{A}; f_\psi) = z_{t+1} - f_\psi(z_{t-H:t}, a_{t-H:t}).
\end{equation}

Optimization proceeds using a standard primal--dual approach~\citep{bertsikas}, alternating between gradient-based updates of the primal variables $(\mathcal{Z}, \mathcal{A})$ and the dual variables $\Lambda$. In particular, we perform $I_{\mathrm{ALM}}$ inner (primal) iterations per outer (dual) step and run the optimization for $O_{\mathrm{ALM}}$ outer iterations using Adam~\citep{kingma2014adam}. We employ a geometric continuation schedule for $\rho$ to progressively enforce dynamic constraints, increasing the penalty at each outer iteration with penalty growth rate $\gamma > 1$ up to a maximum value $\rho_{\max}$. This schedule enables prioritizing video guidance in early iterations, while progressively enforcing physical consistency as $\rho$ increases. For the bounded action constraints, we apply a smooth action reparameterization, following prior work in continuous-control optimization~\citep{haarnoja2018soft}. We empirically validate this smooth reparameterization against projected gradient descent in the ablation study in Appendix~\ref{app:abl}.

\subsubsection{Execution via Model Predictive Control}

The ALM optimization solves for a dynamically feasible latent state--action trajectory $(\mathcal{Z}, \mathcal{A})$ that realizes the latent states of the video plan $z_{\mathrm{vid}}$. The latent state variables $\mathcal{Z}$ are initialized from the latent states of the video plan $z_{\mathrm{vid}}$. This initialization anchors the optimization to a semantic prior, improving the recovery of feasible plans, as shown in Ablation~\ref{sec:abl_main}. When low-level proprioceptive information is available, we initialize the corresponding states using a static prior at the current state $s_t$. Upon convergence, the optimization yields an optimized latent trajectory and action sequence $(\mathcal{Z}^*, \mathcal{A}^*)$, where the action sequence $\mathcal{A}^*$ is executed using a receding-horizon MPC. Conditioning video-guided latent collocation on the current state reduces error accumulation over long horizons, since the current state is enforced through the dynamics constraints in latent-space trajectory optimization. Optionally, we refine the optimized action sequence $\mathcal{A}^\ast$ by sampling $C$ candidate sequences from $\mathcal{N}(\mathcal{A}^\ast, \sigma^2 I)$, evaluating them via open-loop rollouts under the latent dynamics $f_\psi$, and selecting the sequence with the lowest terminal goal cost. Given an execution horizon $K$, GVP-WM executes the first $K$ actions from the optimal trajectory. We analyze the impact of MPC and local refinement via ablations in Appendix~\ref{app:abl}.

\section{Experiments}
We evaluate GVP-WM along three axes: (i) its ability to ground zero-shot video plans into feasible actions under out-of-distribution environments, (ii) robustness to temporally inconsistent video guidance, and (iii) performance in long-horizon settings. We compare against both world-model-based planners without video guidance and direct video-to-action baselines. We further provide ablation studies analyzing the contributions of video guidance and the latent collocation formulation, as well as a qualitative evaluation.

\subsection{Experimental Setup} \label{sec:experiments} 

\subsubsection{Environments and Evaluation Setup} \label{sec:eval_setuo} 

We evaluate GVP-WM in two control environments across multiple long-horizon settings, following the evaluation protocols established in prior work~\citep{dino-wm}. Push-T~\cite{chi2025diffusion} is a contact-rich 2D manipulation task in which an agent pushes a T-shaped object to a target configuration, while Wall is a 2D navigation task that requires visual planning. For both environments, we use DINO-WM~\citep{dino-wm}, a pre-trained action-conditioned world model that leverages DINOv2~\citep{oquab2023dinov2} as a frozen visual encoder. Video plans are generated using  Wan2.1-FLF2V-14B (720p) image-to-video diffusion model~\cite{wan2025} in a zero-shot setting (WAN-0S). Since both environments are out-of-distribution relative to the internet-scale pretraining data of the video model, we additionally fine-tune the video model using $\mathcal{D}_{\text{demo}} = 100$ task-specific demonstrations via LoRA~\citep{hu2022lora}, and generate video plans with domain-adapted video guidance (WAN-FT). In addition, we consider video plans obtained from expert policies that are in-distribution with respect to world-model training (ORACLE). To evaluate robustness to temporally inconsistent video guidance, we introduce synthetic motion blur by temporally averaging $k$ consecutive frames (MB-$k$). In contrast to prior work~\citep{dino-wm}, goal states are specified as visual observations, with no access to proprioceptive goal information during planning. We evaluate planning performance for horizons $T \in {25, 50, 80}$ in Push-T and $T \in {25, 50}$ in Wall, using receding-horizon execution. For each setting, we evaluate on 50 initial–goal pairs sampled from expert policy trajectories of the corresponding horizon length. We report Success Rate, defined as the fraction of successful rollouts across all evaluation episodes. Full implementation details are provided in Appendix~\ref{app:impl_details}.

\subsubsection{Baselines}

We compare against two MPC-based planners that use the same pre-trained action-conditioned world model (DINO-WM), without relying on video guidance. MPC-CEM is a shooting-based method that uses the Cross-Entropy Method to optimize action sequences by sampling candidate trajectories and evaluating them via open-loop rollouts in the latent world model. MPC-GD is a gradient-based planner that optimizes action sequences by backpropagating through the differentiable world-model dynamics. For both methods, planning is performed in the learned latent world model with receding-horizon execution ($K=1$). We also compare against UniPi~\citep{unipi}, a direct video-to-action baseline based on inverse dynamics that maps video trajectories to actions without enforcing world-model dynamics. For a fair comparison, UniPi is provided with the same video plans as GVP-WM; thus, the methods differ only in whether video guidance is executed directly or grounded through a learned world model. Further details are provided in Appendix~\ref{app:impl_details}.

\subsection{Grounding I2V-Generated Video Plans}
\label{sec:main_res}

\begin{table}[htbp]
    \centering
    \caption{
        Success Rate comparison of methods on PushT and Wall for planning horizons $T$. WAN-0S: zero-shot; WAN-FT: domain-adapted; ORACLE: expert (upper bound).
    }
    \label{tab:main-results-pusht-wall-1}
    \vskip 0.05in
    \begin{small}
    \begin{sc}
    \setlength{\tabcolsep}{4pt}
    \renewcommand{\arraystretch}{1.15}
    \begin{tabular}{l|ccc|cc}
    \toprule
     & \multicolumn{3}{c|}{\textbf{PushT}} & \multicolumn{2}{c}{\textbf{Wall}} \\
    \textbf{Method}
    & \textbf{T=25} & \textbf{T=50} & \textbf{T=80}
    & \textbf{T=25} & \textbf{T=50} \\
    \midrule
    MPC-CEM & \underline{0.74} & \underline{0.28} & \textbf{0.06} & \underline{0.92} & 0.74 \\
    MPC-GD  & 0.32 & 0.04 & 0.00 & 0.04 & 0.04 \\
    \midrule
    UniPi (WAN-0S)  & 0.00 & 0.00 & 0.00 & 0.30 & 0.14 \\
    UniPi (WAN-FT) & 0.10 & 0.00 & 0.00 & 0.40 & 0.18 \\
    UniPi (\textit{ORACLE})  & \textit{0.52} & \textit{0.18} & \textit{0.08} & \textit{1.00} & \textit{1.00} \\
    \midrule
    GVP-WM (WAN-0S)  & 0.56 & 0.12 & \underline{0.04} & 0.86 & \underline{0.76} \\
    GVP-WM (WAN-FT) & \textbf{0.80} & \textbf{0.30} & \textbf{0.06} & \textbf{0.94} & \textbf{0.90} \\
    GVP-WM (\textit{ORACLE})  & \textit{0.98} & \textit{0.72} & 0.36 & \textit{1.00} & \textit{1.00} \\
    \bottomrule
    \end{tabular}
    \end{sc}
    \end{small}
    \vskip -0.1in
\end{table}

\begin{table}[htbp]
    \centering
    \caption{
        Average planning time per episode in seconds.
    }
    \label{tab:planning-time-seconds}
    \vskip 0.05in
    \begin{small}
    \begin{sc}
    \setlength{\tabcolsep}{4pt}
    \renewcommand{\arraystretch}{1.15}
    \begin{tabular}{l|ccc|cc}
    \toprule
     & \multicolumn{3}{c|}{\textbf{PushT}} & \multicolumn{2}{c}{\textbf{Wall}} \\
    \textbf{Method}
    & \textbf{T=25} & \textbf{T=50} & \textbf{T=80}
    & \textbf{T=25} & \textbf{T=50} \\
    \midrule
    MPC-CEM & 277 & 596 & 790 & 121 & 198 \\
    MPC-GD  & 96  & 208 & 475 & 44 & 74 \\
    GVP-WM  & \textbf{86} & \textbf{158} & \textbf{208} & \textbf{24} & \textbf{43} \\
    \bottomrule
    \end{tabular}
    \end{sc}
    \end{small}
    \vskip -0.05in
\end{table}

Table~\ref{tab:main-results-pusht-wall-1} reports success rates on Push-T and Wall across increasing planning horizons for all methods and sources of video guidance. We compare GVP-WM against MPC-based planners that do not use video guidance (MPC-CEM, MPC-GD), as well as a direct video-to-action baseline based on inverse dynamics (UniPi). For methods that use video guidance (GVP-WM and UniPi), we consider three sources of video plans: zero-shot generated videos (WAN-0S), domain-adapted generated videos (WAN-FT), and in-distribution expert videos (ORACLE) serving as upper bound. Under domain-adapted video guidance, GVP-WM outperforms MPC-CEM in all evaluated settings except Push-T at $T{=}80$, where performance is comparable. In the zero-shot setting, video guidance is out-of-distribution for the video model in both environments. Consequently, for WAN-0S, MPC-CEM performs better overall, except on Wall at $T{=}50$. However, GVP-WM requires substantially less planning time than MPC-CEM. Under both zero-shot and domain-adapted video guidance, GVP-WM consistently outperforms MPC-GD across environments and planning horizons, particularly on Wall, where gradient-based planning exhibits limitations in visual navigation. In addition, GVP-WM performance with in-distribution expert video plans highlights the potential of video-guided latent collocation to outperform sampling-based and gradient-based planners by grounding video plans in learned world-model dynamics.


GVP-WM outperforms UniPi across both tasks and planning horizons. For Push-T, UniPi fails under zero-shot and domain-adapted video guidance. In contrast, GVP-WM with domain-adapted video guidance achieves strong performance at horizons $T{=}25$ and $50$. Under zero-shot guidance, GVP-WM performance decreases but still recovers feasible plans, indicating robustness. We observe similar trends on Wall under video-generated guidance, with higher overall performance than Push-T. With in-distribution expert videos, GVP-WM outperforms UniPi on Push-T at all horizons. On Wall, UniPi and GVP-WM achieve perfect success provided with oracle videos at both horizons, indicating that inferring actions from video plans is more effective for tasks relying on visual planning than for contact-rich manipulation. In addition, UniPi has negligible inference time, requiring only a single forward pass at test time. We conclude that GVP-WM, unlike UniPi, recovers feasible actions from image-to-video–generated plans across both environments by grounding video plans in world-model dynamics.

\subsection{Robustness to Motion Blur}
\label{sec:res_blur}

\begin{table}[htbp]
    \centering
    \caption{
        Success Rate under increasing levels of motion blur. MB-$k$ denotes temporal averaging over $k$ consecutive frames.
    }
    \label{tab:resblur}
    \vskip 0.05in
    \begin{small}
    \begin{sc}
    \setlength{\tabcolsep}{4pt}
    \renewcommand{\arraystretch}{1.15}
    \begin{tabular}{ll|ccc|cc}
    \toprule
    \textbf{Source} & \textbf{Method}
    & \multicolumn{3}{c|}{\textbf{PushT}}
    & \multicolumn{2}{c}{\textbf{Wall}} \\
     & 
    & \textbf{T=25} & \textbf{T=50} & \textbf{T=80}
    & \textbf{T=25} & \textbf{T=50} \\
    \midrule

    \multirow{2}{*}{MB-10}
    & UniPi & 0.03 & 0.00 & 0.00 & 0.00 & 0.02 \\
    & GVP-WM & \textbf{0.82} & \textbf{0.46} & \textbf{0.08} & \textbf{0.94} & \textbf{1.00} \\
    \midrule

    \multirow{2}{*}{MB-5}
    & UniPi & 0.04 & 0.00 & 0.06 & 0.40 & 0.52 \\
    & GVP-WM & \textbf{0.94} & \textbf{0.54} & \textbf{0.16} & \textbf{1.00} & \textbf{1.00} \\
    \midrule

    \multirow{2}{*}{MB-3}
    & UniPi & 0.16 & 0.00 & 0.00 & 0.52 & 0.82 \\
    & GVP-WM & \textbf{0.96} & \textbf{0.70} & \textbf{0.34} & \textbf{1.00} & \textbf{1.00} \\
    \midrule

    \multirow{2}{*}{Oracle}
    & UniPi & 0.52 & 0.18 & 0.08 & \textbf{1.00} & \textbf{1.00} \\
    & GVP-WM & \textbf{0.98} & \textbf{0.72} & \textbf{0.36} & \textbf{1.00} & \textbf{1.00} \\
    \bottomrule
    \end{tabular}
    \end{sc}
    \end{small}
    \label{table:esblur}
    \vskip -0.1in
\end{table}


To evaluate robustness to temporally inconsistent video guidance, we introduce synthetic motion blur into expert videos by temporally averaging consecutive frames over sliding windows of size 3, 5, and 10 (MB-3, MB-5, MB-10). Table~\ref{table:esblur} reports success rates for UniPi and GVP-WM under increasing levels of temporal blur. UniPi is highly sensitive to temporally degraded guidance. At horizon $T{=}25$, success drops from $0.52$ under clean expert videos to $0.16$, $0.04$, and $0.02$ under MB-3, MB-5, and MB-10, respectively. At $T{=}50$, UniPi achieves only $0.18$ success under clean videos and fails entirely under any level of blur. In contrast, GVP-WM remains robust to temporal inconsistencies in video guidance. At $T{=}25$, it maintains high success under mild and moderate blur ($0.96$ and $0.94$) and retains strong performance even under severe blur ($0.82$). At $T{=}50$, performance degrades gradually as blur increases, from $0.72$ under clean videos to $0.70$, $0.54$, and $0.46$ under MB-3, MB-5, and MB-10. Across all settings, GVP-WM substantially outperforms UniPi, with the performance gap widening at longer horizons and under stronger temporal blur. These results show that GVP-WM remains effective when video guidance exhibits temporal inconsistencies that violate physical constraints, whereas inverse-dynamics models fail under such conditions.

\subsection{Qualitative Analysis of I2V-generated Video Plans}
\label{seq:qual}
\begin{figure*}[t]
    \centering
    
    \begin{subfigure}[t]{0.48\textwidth}
        \centering
        \includegraphics[width=\linewidth]{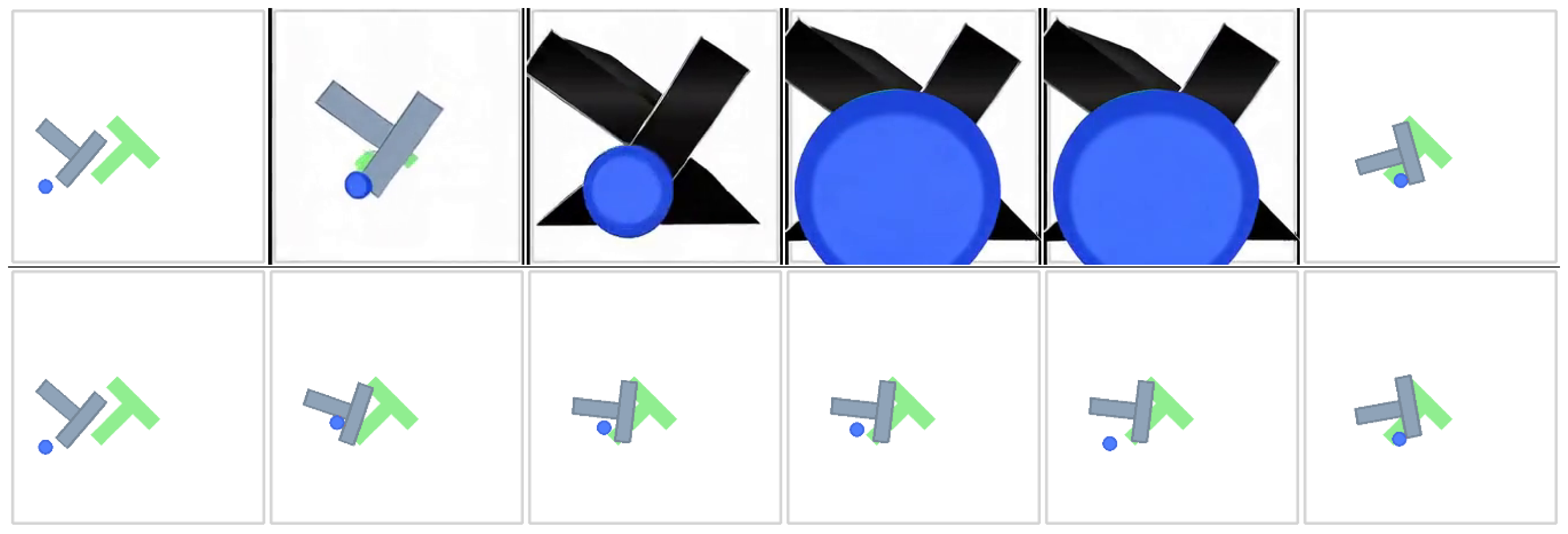}
        \caption{Morphological Drift (Success)} 
        \label{fig:ood-circle-success}
    \end{subfigure}
    \hfill
    \begin{subfigure}[t]{0.48\textwidth}
        \centering
        \includegraphics[width=\linewidth]{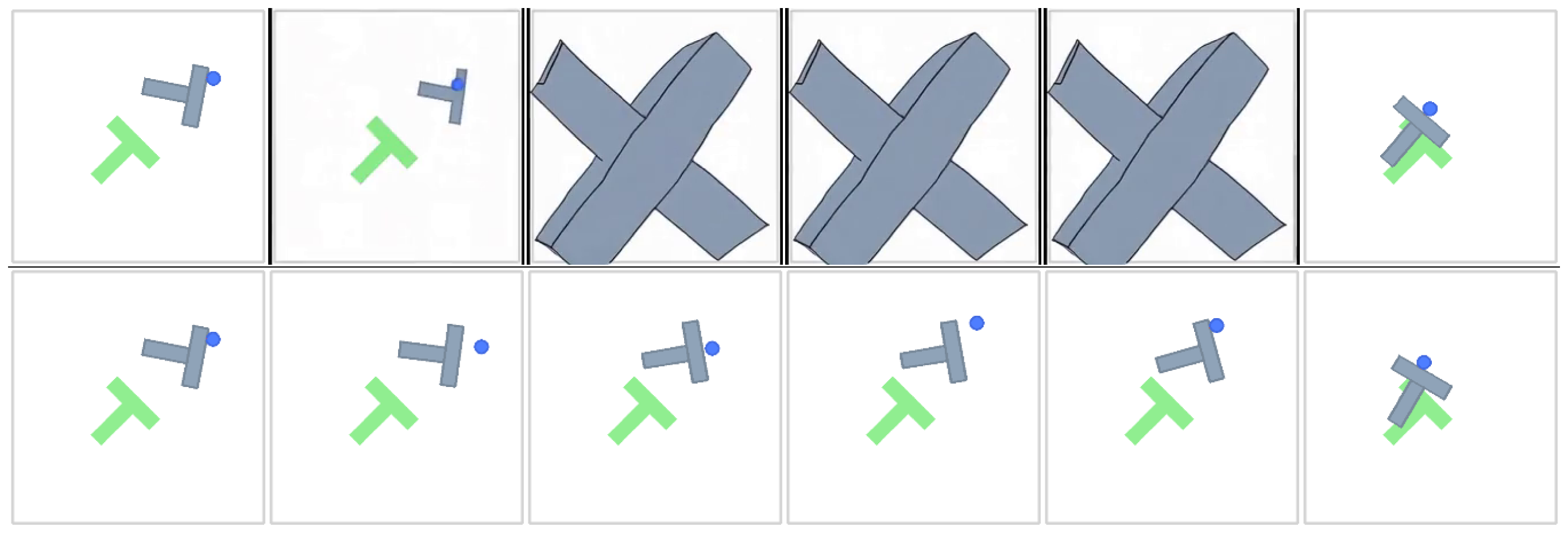}
        \caption{Morphological Drift (Failure)} 
        \label{fig:ood-cross-fail}
    \end{subfigure}

    \vspace{0.2cm} 

    \begin{subfigure}[t]{0.48\textwidth}
        \centering
        \includegraphics[width=\linewidth]{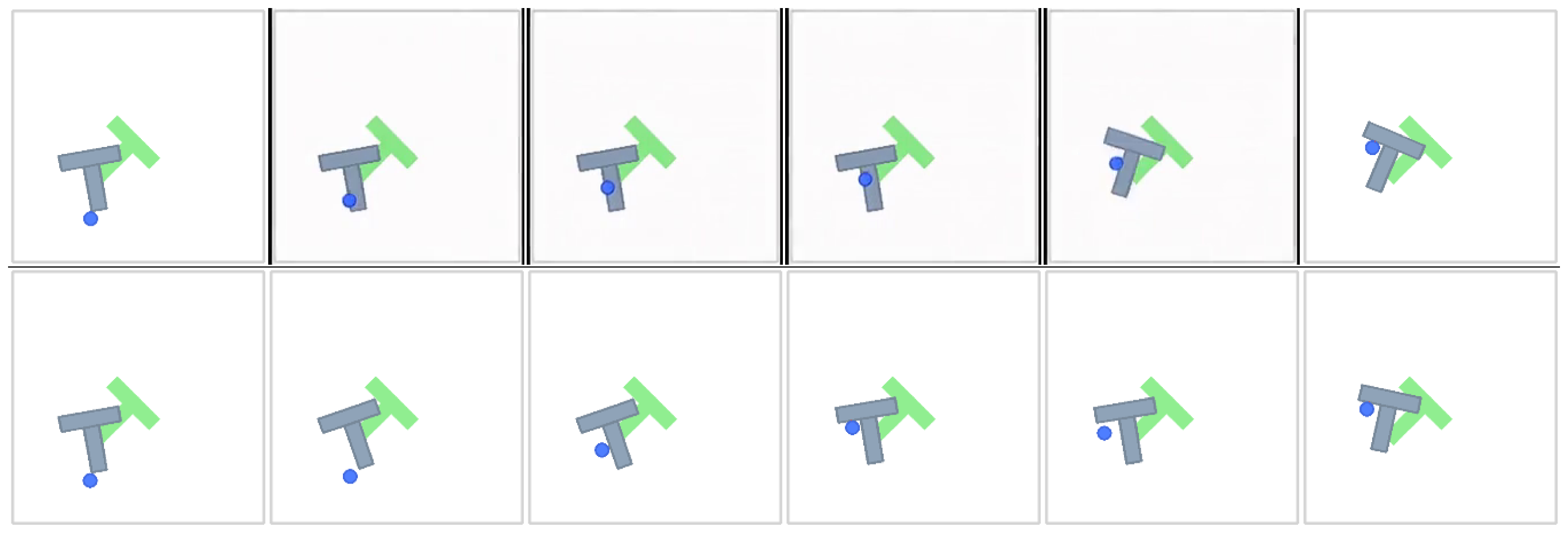}
        \caption{Rigid-Object Physics Violation (Failure)} 
        \label{fig:physics-violation-main}
    \end{subfigure}
    \hfill
    \begin{subfigure}[t]{0.48\textwidth}
        \centering
        \includegraphics[width=\linewidth]{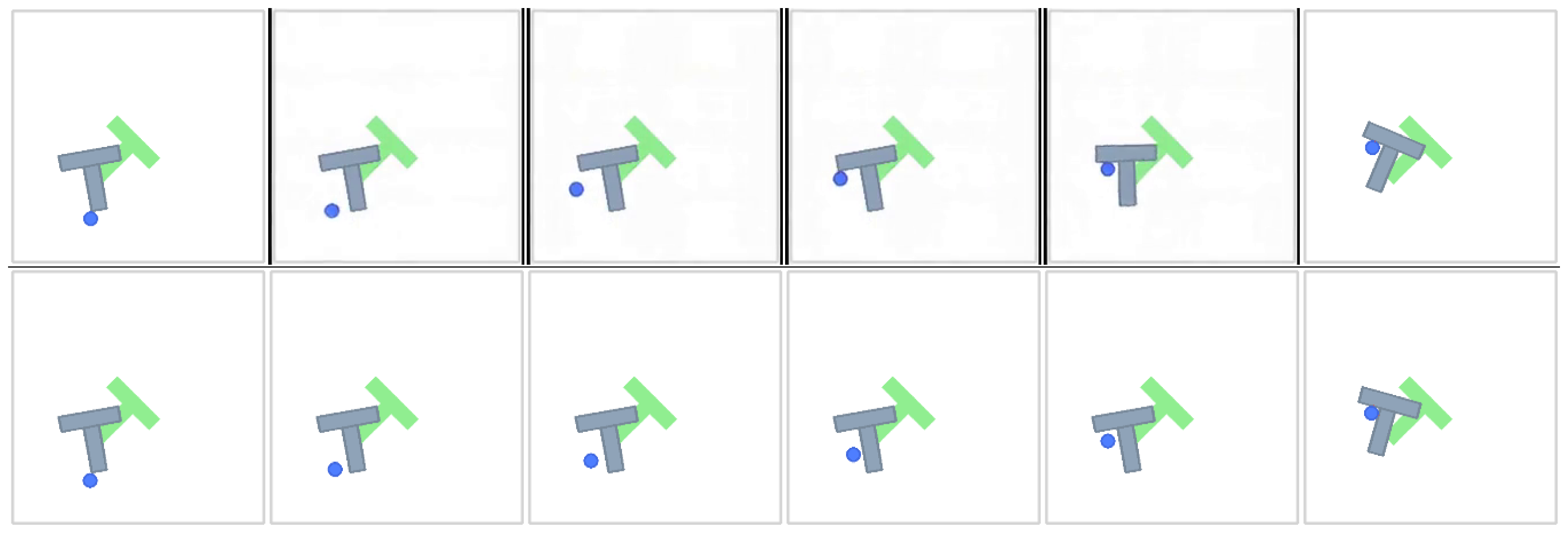}
        \caption{ Domain-Adapted Video Guidance (Success)} 
        \label{fig:teleportation}
    \end{subfigure}

     \vspace{0.2cm} 

    \begin{subfigure}[t]{0.48\textwidth}
        \centering
        \includegraphics[width=\linewidth]{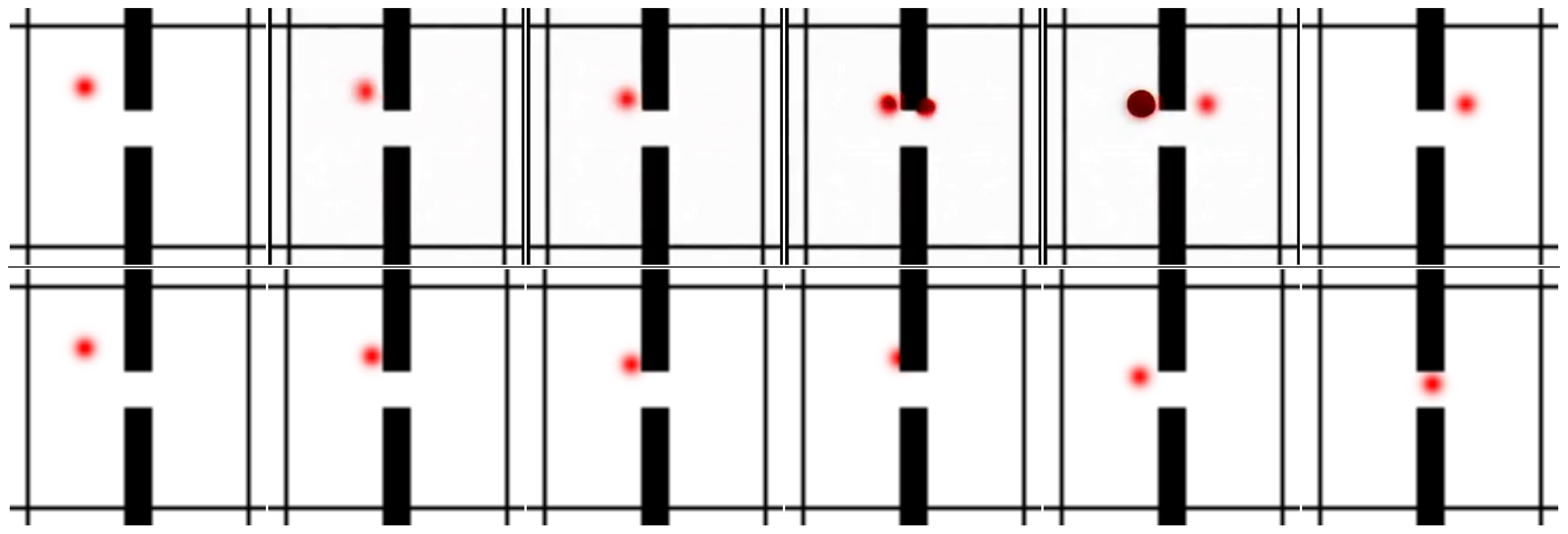}
        \caption{Spatial Bilocation (Failure)} 
        \label{fig:bilocation-f}
    \end{subfigure}
    \hfill
    \begin{subfigure}[t]{0.48\textwidth}
        \centering
        \includegraphics[width=\linewidth]{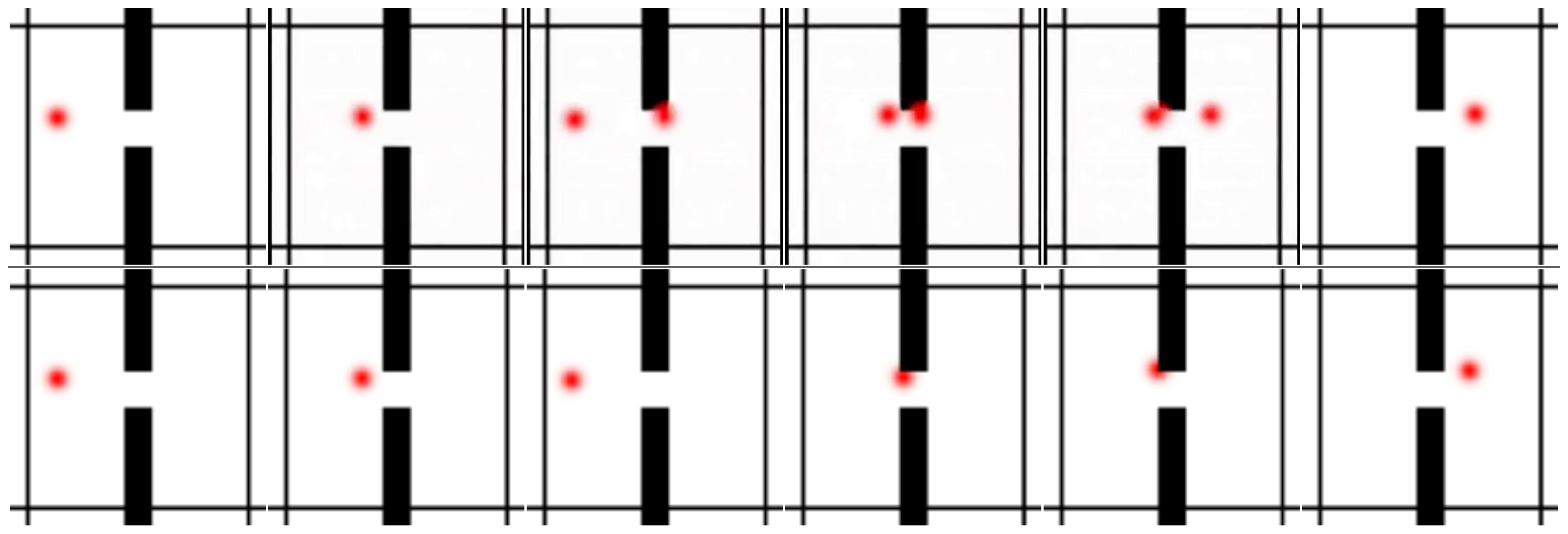}
        \caption{ Spatial Bilocation (Success)} 
        \label{fig:bilocation-s}
    \end{subfigure}
    \vspace{0.2cm}

     \caption{
Qualitative comparison of video-generated plans (top) and GVP-WM executions (bottom). In Push-T, zero-shot video plans (WAN-0S) exhibit physical inconsistencies, including morphological drift (a–b) and rigid-object physics violations (c), while domain-adapted video guidance (WAN-FT) produces physically consistent plans. In Wall, zero-shot video plans exhibit spatial bilocation (e-f).
}    
    \label{fig:wan-qualitative-analysis}
\end{figure*}

Figure~\ref{fig:wan-qualitative-analysis} compares image-to-video (I2V) plans generated by the Wan2.1-FLF2V-14B diffusion model with execution trajectories produced by GVP-WM. In our experimental setup, zero-shot I2V-generated plans frequently violate physical constraints, as both Push-T and Wall simulation environments are out of distribution relative to the internet-scale pretraining data of the video model. In Push-T, zero-shot plans often exhibit morphological drift, causing the appearance or identity of the manipulated object to change across frames. GVP-WM can sometimes recover a feasible trajectory by enforcing world-model dynamics and disregarding inconsistent video frames (Figure~\ref{fig:ood-circle-success}); however, severe semantic drift can still lead to execution failure (Figure~\ref{fig:ood-cross-fail}). In addition, zero-shot video plans can violate rigid-object physics (Figure~\ref{fig:physics-violation-main}). In contrast, domain-adapted video guidance from a diffusion model fine-tuned with LoRA on a small dataset (Section~\ref{sec:eval_setuo}) produces more physically consistent plans, which GVP-WM is able to follow to successfully complete the task (Figure~\ref{fig:teleportation}). In Wall, zero-shot video plans can additionally exhibit spatial bilocation, where the agent appears at multiple spatial locations within the same frame; despite this violation, GVP-WM is still able to recover feasible execution trajectories in some cases (Figure~\ref{fig:bilocation-s}). Additional qualitative examples and failure modes across both environments are provided in Appendix~\ref{app:qual_eval}.

\subsection{Ablation Studies}
\label{sec:abl_main} 

\begin{table}[htbp]
    \centering
    \caption{
        Ablation study on Push-T ($T=25$). We evaluate the impact of video guidance and latent collocation.
    }
    \label{tab:ablation}
    \vskip 0.05in
    \begin{small}
    \begin{sc}
    \setlength{\tabcolsep}{5pt}
    \renewcommand{\arraystretch}{1.15}
    \begin{tabular}{lccc}
    \toprule
    \textbf{Method} & \textbf{WAN-0S} & \textbf{WAN-FT} & \textbf{ORACLE} \\
    \midrule
    No Video Guidance  & \textbf{0.68} & 0.68   & 0.68   \\
    No Video Init  & 0.66 & 0.60 & 0.62 \\
    No Video Loss  & 0.54 & 0.72 & 0.92 \\
    No Collocation & 0.08 & 0.12 & 0.60 \\
    \midrule
    \textbf{GVP-WM} & 0.56 & \textbf{0.82} & \textbf{0.98} \\
    \bottomrule
    \end{tabular}
    \end{sc}
    \end{small}
    \vskip -0.05in
\end{table}

We study the impact of video guidance and latent collocation; additional ablations on GVP-WM design choices are presented in Appendix~\ref{app:abl}. Removing video guidance by using random latent state initialization with $\lambda_v = 0$ affects performance differently depending on video quality. Specifically, it improves performance under zero-shot guidance (WAN-0S), but degrades performance under higher-quality guidance, with success decreasing from $0.82$ to $0.68$ for domain-adapted videos (WAN-FT) and from $0.98$ to $0.68$ for oracle video plans. Removing latent state initialization causes a substantial performance drop under both domain-adapted and oracle guidance, with success decreasing from $0.82$ to $0.60$ under WAN-FT and from $0.98$ to $0.62$ under oracle videos, highlighting the importance of video-based initialization. Including the video alignment loss improves performance across all video sources, with particularly strong gains under domain-adapted guidance (from $0.72$ to $0.82$), demonstrating the contribution of video guidance during latent collocation. In addition, we study the impact of latent collocation by fixing the latent states to the video plan $z_{0:T} = z^{vid}_{0:T}$  and optimizing only the actions $a_{0:T}$. Performance collapses to $0.12$ under domain-adapted video guidance, demonstrating that latent trajectories produced by video models are not dynamically feasible and cannot be used directly as waypoints. This behavior is consistent with the failures observed for inverse-dynamics baselines (Table~\ref{tab:main-results-pusht-wall-1}) and highlights the necessity of jointly optimizing latent states and actions under world-model dynamics.

\section{Related Work}

\subsection{Video Models for Decision Making}

Large-scale generative video models~\citep{ho2022video, blattmann2023stable} have been widely explored as zero-shot visual planners. LVP~\citep{largevideoplans} proposes a foundation-scale video generative model for robotics, leveraging diffusion history guidance~\citep{song2025history} and diffusion forcing~\citep{chen2024diffusionforcing} to improve temporal coherence. Other approaches enhance video generation by conditioning on pose~\citep{bai2025peva}, motion~\citep{hu2023animate}, or physical constraints~\citep{yuan2023physdiff}. Beyond video generation, several works infer actions directly from generated videos~\citep{baker2022video, unipi, ko2023learning}. UniPi~\citep{unipi} synthesizes expert demonstrations via text-conditioned video generation and maps video frames to actions using an inverse dynamics model. However, despite progress in inferring actions from videos~\citep{edwards2019imitating,pavlakos2024reconstructing,potamias2025wilor,li2025megasam,zhang2025latent}, mapping video-generated plans to physically feasible action sequences remains a challenge~\citep{largevideoplans,ni2025gemmimic}. In contrast, hierarchical approaches treat generated videos as high-level guidance rather than direct action supervision. HVF~\citep{nair2020hierarchical} generates visual subgoals via video prediction and connects them using model-predictive control, while VLP~\citep{du2023video} performs tree search over generated video futures to solve long-horizon tasks. SuSIE~\citep{black2023zero} leverages image-editing models to synthesize sparse visual subgoals that guide a low-level, goal-conditioned policy. However, these approaches generally assume that generated videos or visual subgoals are physically feasible in the target environment. Recent work~\citep{luo2025grounding} addresses this limitation by using generated videos as training-time guidance during policy learning. In contrast, our method grounds a generated video plan at inference time by directly optimizing latent states and actions under a learned, action-conditioned world model, enabling test-time grounding without additional environment interaction or policy training. 

\subsection{Planning with World Models}

World models have been widely used for learning environment dynamics across domains such as robotics~\citep{yang2023unisim, nvidia2025cosmos}, autonomous driving~\citep{hu2023gaia1, russell2025gaia}, and video games~\citep{bruce2024genie, hafner2025dreamer4}. These models learn latent representations of environment dynamics conditioned on actions, supporting planning~\citep{lecun2022path, planet} and model-based reinforcement learning through rollouts in imagination~\citep{hafner2023mastering, hafner2025dreamer4}. Despite progress in goal-conditioned universal value function learning~\citep{ma2024vision,ziakasvita}, policy learning in world models may fail to generalize due to reliance on zero-shot reward models. In contrast, planning enables inference-time optimization of action sequences via simulated rollouts under world-model dynamics, without requiring policy learning or reward functions~\citep{dreamer,lecun2022path}. Shooting-based methods perform trajectory optimization by evaluating candidate action sequences~\citep{dino-wm, bar2025navigation}, while gradient-based methods directly optimize trajectories through the learned dynamics~\citep{terver2026what, dino-wm}. Latent collocation has been proposed as a formulation that optimizes over latent state trajectories while enforcing dynamics consistency~\citep{latco}. In this work, we propose video-guided collocation, which grounds visual plans in a learned world model and improves performance when sufficient video guidance is available. By anchoring latent trajectories to a semantic prior, GVP-WM also reduces the effective search space, leading to faster convergence.

\section{Discussion, Limitation, and Future Work}
\subsection{Discussion}

GVP-WM grounds video-generated plans into physically feasible action sequences using action-conditioned world models. Large-scale video models, while powerful, frequently violate low-level physical constraints when applied to out-of-distribution robotic environments such as Push-T, producing artifacts including object teleportation, rigid-body violations, and temporal inconsistencies. In such cases, directly mapping video frames to actions via inverse-dynamics models fails because the underlying visual plans are not dynamically feasible. Our empirical results show that GVP-WM can recover executable action sequences even when video plans are physically invalid or temporally inconsistent, and that it consistently outperforms inverse-dynamics baselines across environments and planning horizons. Performance under domain-adapted video guidance and in-distribution oracle video plans further indicates the potential of video-guided planning. These results suggest that as large-scale video models and action-conditioned world models continue to improve, GVP-WM can provide a robust mechanism for mapping video plans into executable actions.

\subsection{Limitations and Future Work}
GVP-WM relies on learned world-model dynamics; in zero-shot settings, mismatches between the learned dynamics and the environment can lead to infeasible plans regardless of video quality. Although faster than sampling-based planners, GVP-WM still requires iterative test-time optimization, which may limit applicability in real-world settings. While GVP-WM is robust to motion blur, severely misaligned zero-shot video guidance can degrade performance relative to unguided planning. Future work includes extending GVP-WM to real-world robotic manipulation, where zero-shot video guidance is more likely to be in-distribution, as well as exploring hierarchical planning and policy distillation.

\section*{Acknowledgments}

We thank Lennart Bastian, Joey Bose, and Tony Danjun Wang for their helpful comments on an earlier draft. This work was supported by UKRI (EP/Y037111/1) as part of the ProSafe project (EU Horizon 2020, MSCA, grant no.\ 101119358).

\newpage
\clearpage

\bibliographystyle{unsrtnat}

\bibliography{example_paper_arxiv}

\newpage
\appendix
\onecolumn

\section{Qualitative Evaluation and Failure Modes}
\label{app:qual_eval}

\begin{figure}[htbp!] 
    \centering
    
    \includegraphics[width=\textwidth]{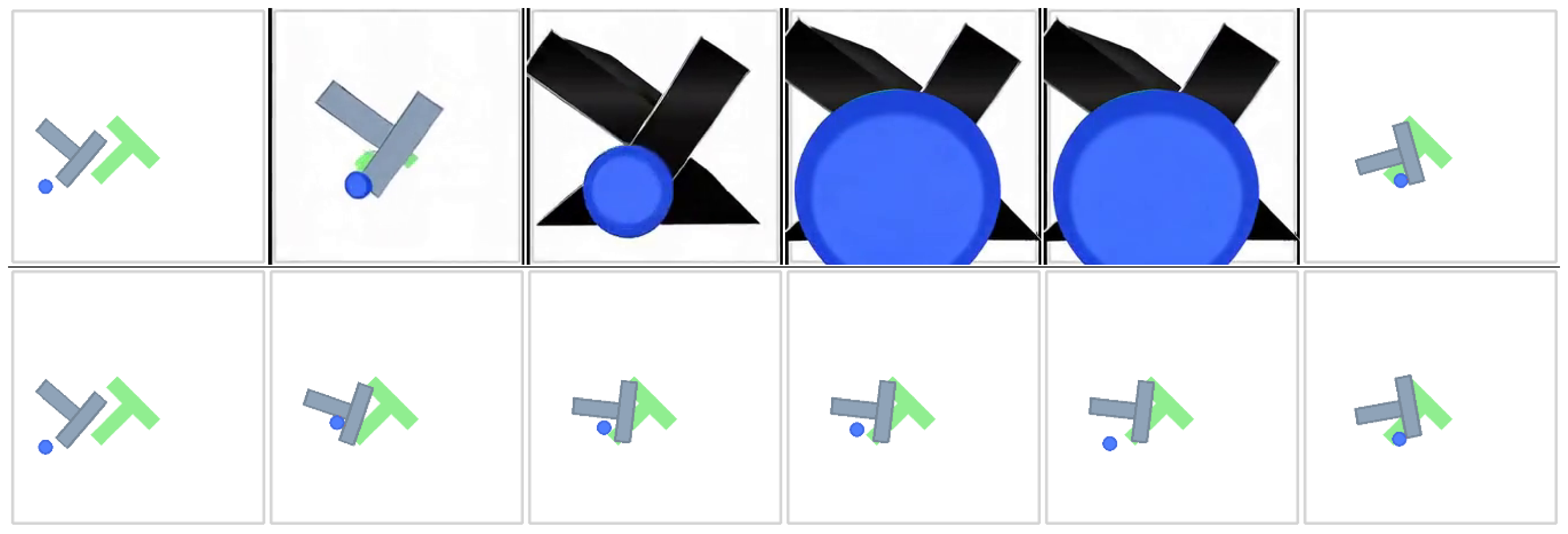}
    \caption{The generated video plan (top) hallucinates large artifacts. GVP-WM (bottom) succeeds.}
    \label{fig:wan-physics-violation-1}
    
    \vspace{1em} 
    
    \includegraphics[width=\textwidth]{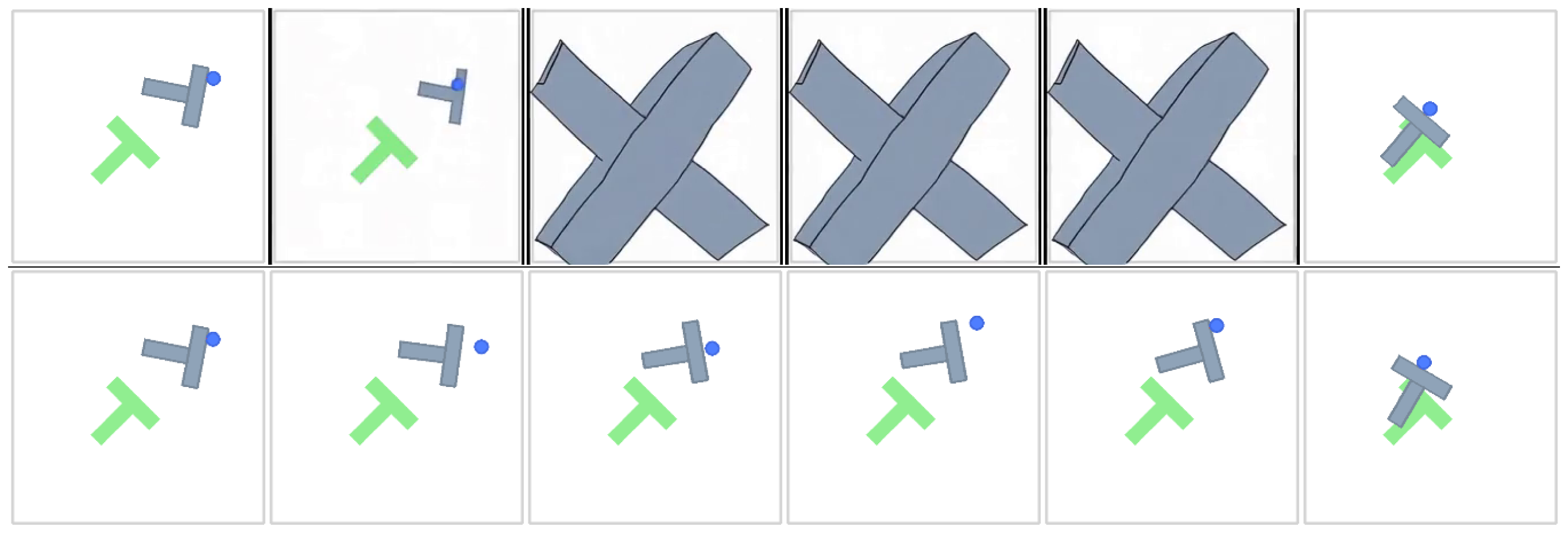}
    \caption{The generated video plan (top) hallucinates large artifacts. GVP-WM (bottom) fails.}
    \label{fig:wan-physics-violation-2}
    
    \vspace{1em}
    
    \includegraphics[width=\textwidth]{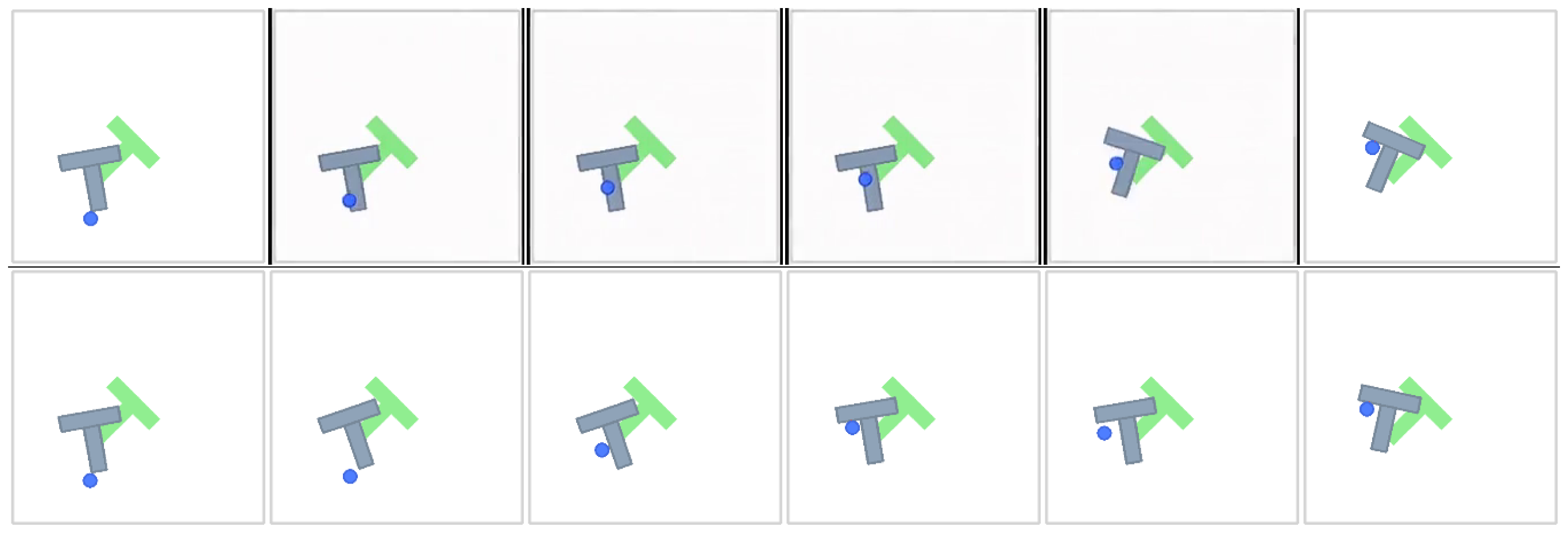}
    \caption{Zero-shot generated video plan (top) violates rigid object physics by passing the agent through T. GVP-WM (bottom) fails.}
    \label{fig:wan-physics-violation-3}
\end{figure}

\clearpage 

\begin{figure}[t!] 
    \centering
    
    \includegraphics[width=\textwidth]{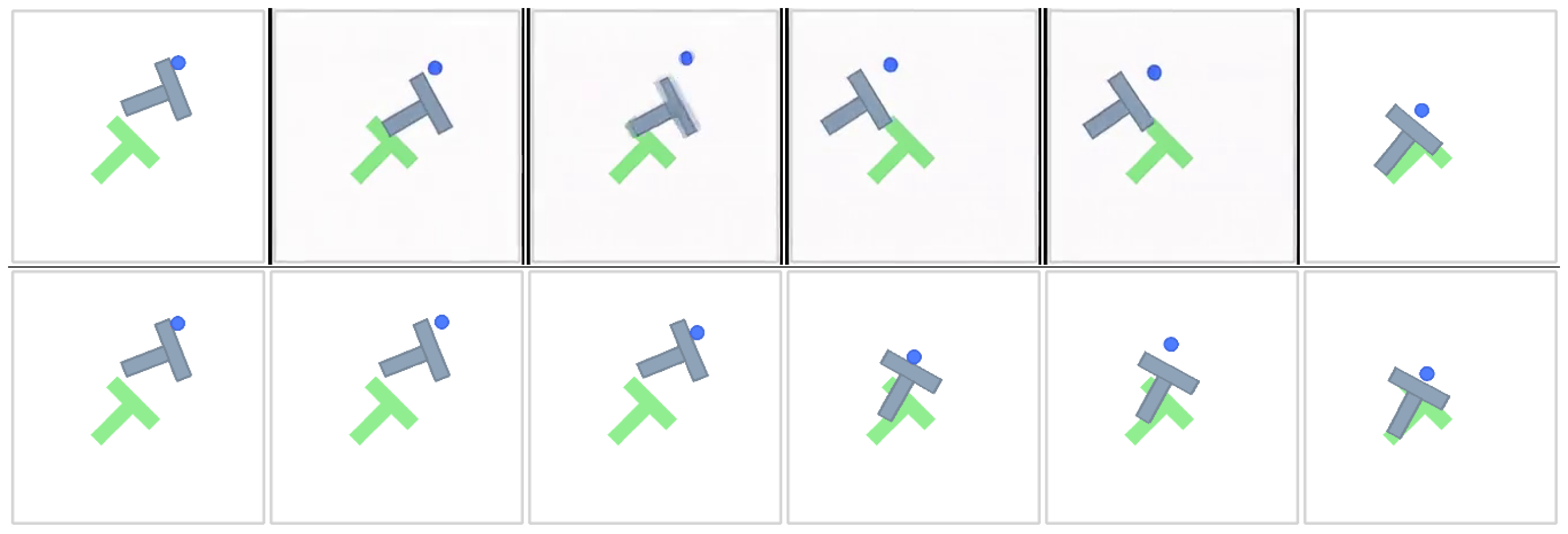}
    \caption{Zero-shot generated video plan (top) violates kinematics by teleporting the agent to goal state. GVP-WM (bottom) fails.}
    \label{fig:wan-physics-violation-4}
    
    \vspace{1em}
    
    \includegraphics[width=\textwidth]{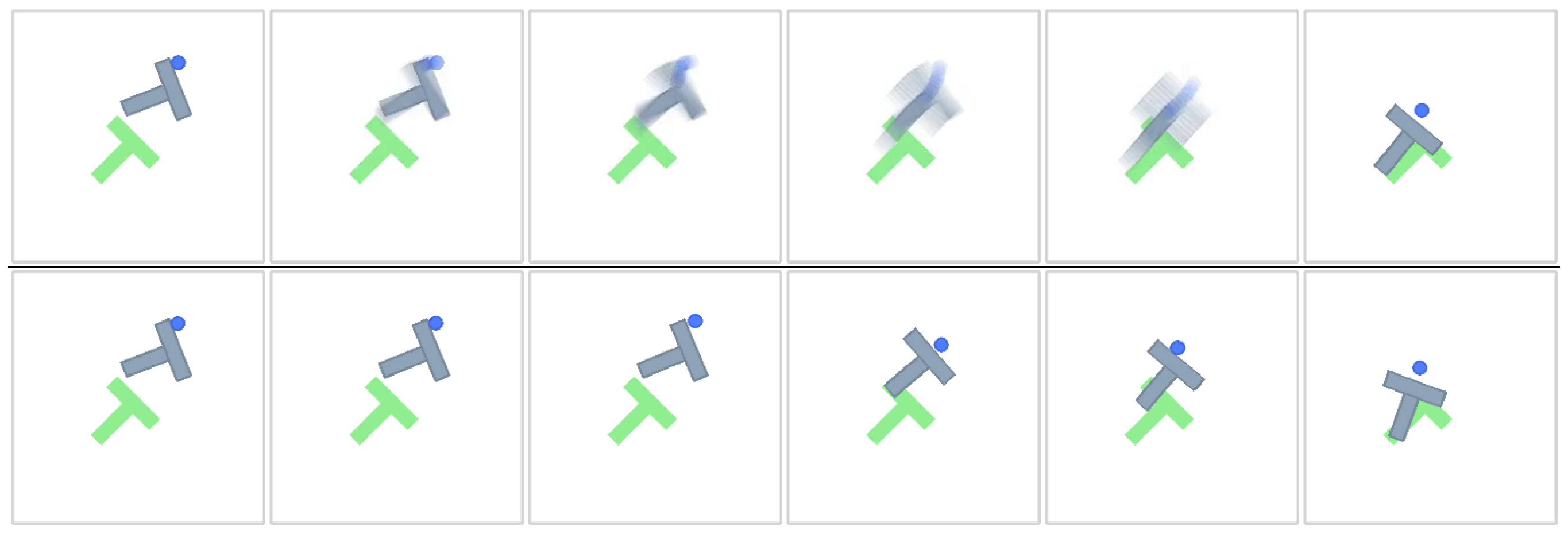}
    \caption{Motion blurred oracle video plan (MB-10) (top). GVP-WM (bottom) fails.}
    \label{fig:blur-robustness}
    
    \vspace{1em}
    
    \includegraphics[width=\textwidth]{figures/rollouts/lora_2row.pdf}
    \caption{Fine-tuned video model (WAN-FT) generates a physically consistent plan (top). GVP-WM (bottom) succeeds.}
    \label{fig:lora-success}
\end{figure}

\begin{figure}[t!] 
    \centering
    \vspace{1em}
    \includegraphics[width=\textwidth]{figures/rollouts/wall_base_failure_2row.pdf}
    \caption{Zero-shot generated video plan (top) exhibits spatial bilocation, depicting the agent at 2 locations. GVP-WM (bottom) fails.}
    \label{fig:wall-basefailue}

    \vspace{1em}
    \includegraphics[width=\textwidth]{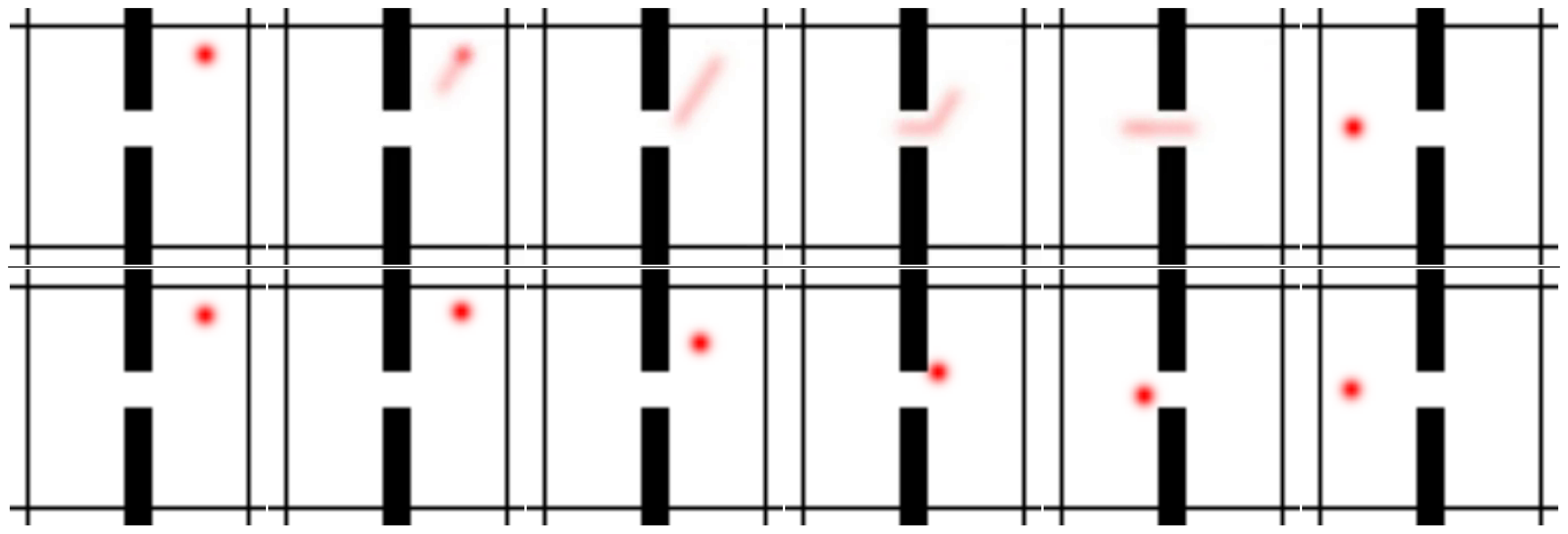}
    \caption{Motion blurred oracle video plan (MB-10) (top). GVP-WM (bottom) succeeds.}
    \label{fig:wall-blursuccess}

    \vspace{1em}
    \includegraphics[width=\textwidth]{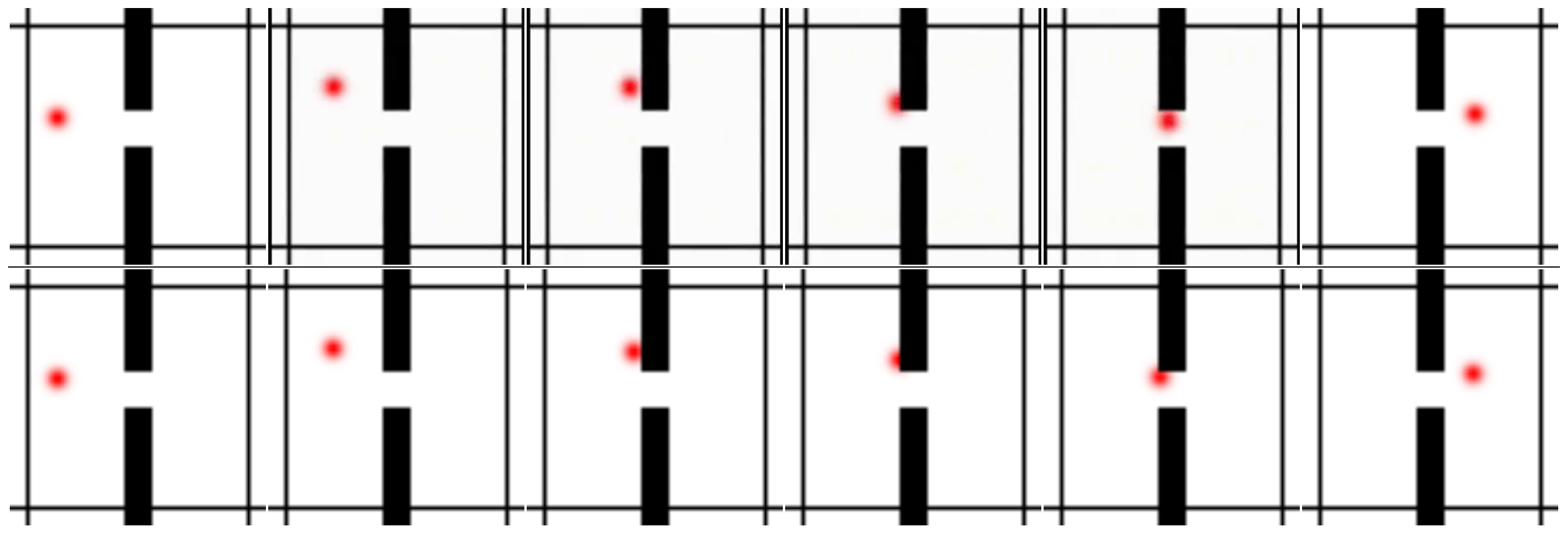}
    \caption{Fine-tuned video model (WAN-FT) generates a physically consistent plan (top). GVP-WM (bottom) succeeds.}
    \label{fig:wall-lorasucess}
\end{figure}

\clearpage
\section{Ablation Studies}
\label{app:abl}
\begin{table}[htbp]
    \centering
    \caption{
        Ablation study on Push-T ($T=25$). We evaluate the contribution of each component by removing or replacing it.
        WAN-0S: Zero-shot video, WAN-FT: Fine-tuned video, ORACLE: Ground-truth video.
    }
    \label{tab:ablation}
    \vskip 0.05in
    \begin{small}
    \begin{sc}
    \setlength{\tabcolsep}{5pt}
    \renewcommand{\arraystretch}{1.15}
    \begin{tabular}{lccc}
    \toprule
    \textbf{Method} & \textbf{WAN-0S} & \textbf{WAN-FT} & \textbf{ORACLE} \\
    \midrule
    \multicolumn{4}{l}{\textit{Video Guidance}} \\
    No Video Guidance  & \textbf{0.68} & 0.68   & 0.68   \\
    No Video Init  & 0.66 & 0.60 & 0.62 \\
    No Video Loss  & 0.54 & 0.72 & 0.92 \\
    \midrule
    \multicolumn{4}{l}{\textit{Collocation \& ALM Solver}} \\
    No Collocation & 0.08 & 0.12 & 0.60 \\
    Projected SGD   & 0.54 & 0.72 & 0.92 \\
    \midrule
    \multicolumn{4}{l}{\textit{Planning \& Execution}} \\
    Open Loop (No MPC)     & 0.44 & 0.80 & 0.86 \\
    No Local Refinement     & 0.52 & 0.72 & 0.92 \\
    \midrule
    \multicolumn{4}{l}{\textit{Video Alignment Metric}} \\
    MSE Alignment     & 0.54 & 0.64 & 0.90 \\
    \midrule
    \textbf{GVP-WM (Ours)} & 0.56 & \textbf{0.82} & \textbf{0.98} \\
    \bottomrule
    \end{tabular}
    \end{sc}
    \end{small}
    \vskip -0.05in
\end{table}

\subsection{Effect of Video Guidance}

In GVP-WM, video plans are used in two ways: (i) to initialize the latent trajectory and (ii) to provide a scale-invariant video alignment loss weighted by $\lambda_v$. Removing video guidance entirely (random latent state initialization with $\lambda_v = 0$) affects performance differently depending on video quality. While removing video guidance improves performance under zero-shot out-of-distribution video plans (WAN-0S), it degrades performance under higher-quality guidance, with success decreasing from $0.82$ to $0.68$ for domain-adapted generated videos (WAN-FT) and from $0.98$ to $0.68$ for oracle video plans.
Removing latent state initialization from the video plan causes a substantial drop under both domain-adapted and oracle guidance: success decreases from $0.82$ to $0.60$ with WAN-FT guidance and from $0.98$ to $0.62$ with oracle video plans, indicating the importance of latent trajectory initialization from video plans.  The video alignment loss consistently improves performance across all video sources. Compared to removing the alignment term ($\lambda_v=0$), adding scale-invariant alignment improves success from $0.54$ to $0.56$ under zero-shot guidance (WAN-0S), from $0.72$ to $0.82$ under domain-adapted guidance (WAN-FT), and from $0.92$ to $0.98$ under oracle video plans. These gains suggest that alignment is beneficial across settings, with larger improvements when the video guidance is more reliable.

\subsection{Effect of Latent Collocation and ALM Solver}

We evaluate the role of latent collocation by fixing latent states to the video plan ($z_t = z_t^{\text{vid}}$) and optimizing only actions.
This ablation causes performance to collapse to $0.12$ under fine-tuned video guidance, demonstrating that latent trajectories produced by video models are not dynamically feasible and cannot be executed directly. We additionally replace our smooth \texttt{tanh} action parameterization with projected gradient descent using action clipping.
This substitution reduces success from $0.82$ to $0.72$, indicating that smooth unconstrained optimization improves convergence of the ALM solver.

\subsection{Effect of MPC Execution}
To study the execution strategy, we remove receding-horizon control and execute the optimized trajectory in open loop. This reduces performance from $0.82$ to $0.80$, showing that feedback during execution helps correct model mismatch. Removing the local refinement step further degrades success to $0.72$, indicating that this lightweight post-optimization refinement is important for resolving residual inaccuracies left by the non-convex objective.

\subsection{Effect of Scale-Invariant Alignment}

Finally, we ablate the video alignment objective by replacing the scale-invariant normalized loss with a standard MSE loss on latent states.
This reduces success under fine-tuned video guidance from $0.82$ to $0.64$, confirming that latent representations from video models exhibit magnitude drift relative to the world model training distribution and that scale invariance is essential for effective grounding.

\section{Implementation Details}
\label{app:impl_details}

\subsection{Environments and Evaluation Protocols}
\label{app:env}

We evaluate in two control environments that are out-of-distribution relative to the internet-scale pretraining data of the video generator. In both environments, goal states are specified purely as visual observations, and no proprioceptive goal information is available during planning. 

\paragraph{Push-T.}
Push-T~\citep{chi2025diffusion} is a contact-rich manipulation task in which a circular agent pushes a T-shaped block toward a target configuration in 2D, requiring reasoning about low-level physical interactions and rigid-body contact dynamics. A rollout is considered successful if the final block pose satisfies both positional and rotational accuracy thresholds relative to the target. We evaluate on 50 distinct initial–goal pairs for planning horizons $T \in \{25, 50, 80\}$, where target goals are drawn from held-out expert demonstrations of corresponding lengths. We report Success Rate (SR), defined as the fraction of successful rollouts across all evaluation episodes.

\paragraph{Wall.}
Wall is a navigation task in which an agent must maneuver around a barrier to reach a target region using abstract 2D visual observations. The task does not involve contact dynamics and primarily tests visual planning and geometric feasibility. Success is defined by the Euclidean distance between the agent’s final 2D position and the target location. We evaluate on 50 initial–goal pairs for planning horizons $T \in \{25, 50\}$, using expert demonstrations that traverse the wall in the corresponding number of steps. We report Success Rate (SR), defined as the fraction of successful rollouts across all evaluation episodes.

\subsection{World Model}
\label{app:world-model}

\begin{table}[t]
\centering
\caption{Dino-WM World Model Architecture and Training Configuration}
\label{tab:world-model}
\begin{tabular}{lll}
\toprule
\textbf{Category} & \textbf{Parameter} & \textbf{Value} \\
\midrule
\multirow{7}{*}{Architecture}
& Visual Encoder & DINOv2-ViT-S/14 (frozen) \\
& Latent Dynamics Predictor & ViT (6 layers, 16 heads, MLP dim 2048) \\
& Decoder (visualization only) & VQ-VAE (384 channels, 2048 entries) \\
& Image Resolution & $224 \times 224$ \\
& History Frames & 3 \\
& Prediction Horizon & 1 frame \\
& Frame Skip & 5 \\
\midrule
\multirow{6}{*}{Training}
& Training Trajectories & 18{,}500 (Push-T) \& 1{,}920 (Wall) \\
& Batch Size & 32 \\
& Training Epochs & 100 \\
& Encoder Learning Rate & $1 \times 10^{-6}$ \\
& Predictor Learning Rate & $5 \times 10^{-4}$ \\
& Decoder / Action Encoder LR & $3 \times 10^{-4}$ / $5 \times 10^{-4}$ \\
\bottomrule
\end{tabular}
\end{table}

We perform planning using pre-trained, action-conditioned world models from \textsc{DINO-WM} \citep{dino-wm}. Separate world models are trained for Push-T and Wall, each using data collected within its respective environment. In Push-T, the world model is trained on expert demonstration trajectories, while in Wall the world model is trained from randomly collected trajectories, following the DINO-WM architecture and training. The world model consists of a frozen DINOv2 ViT-S/14 visual encoder, a transformer-based latent dynamics predictor conditioned on actions, and a VQ-VAE decoder used only for visualization. We use the publicly released DINO-WM training setup and checkpoints without modification. The world models are trained offline and are not fine-tuned during evaluation, isolating the contribution of the planning method. The world model architecture and training configuration are summarized in Table~\ref{tab:world-model}.

\subsection{Video Generation Model}
\label{app:video-model}

We use the Wan2.1-FLF2V-14B (720p) \cite{wan2025} video generation model, conditioned on the first frame, to synthesize open-loop video plans. To bridge the domain gap between internet-scale video data and robotic environments, we apply supervised fine-tuning (SFT) using Low-Rank Adaptation (LoRA) via the DiffSynth framework. Fine-tuning is performed with rank $r = 32$ for 10 epochs, repeating the dataset 10 times per epoch to reduce data loading overhead. Training uses 4 $\times$ NVIDIA A100 (80GB) GPUs, a learning rate of $1 \times 10^{-5}$, and mixed-precision, taking approximately 5 hours and 40 minutes per epoch.

\subsection{Spatial and Temporal Alignment}

The Push-T environment provides $224 \times 224$ square observations, while the Wan2.1 video model expects $1280 \times 720$ inputs with a $16{:}9$ aspect ratio. We apply symmetric padding to the initial and goal observations prior to video generation and crop generated frames before world-model encoding. Since the video generator and world model operate at different temporal strides, generated video latents are interpolated or subsampled to match the world-model frame skip.

\subsection{Configuration}
Planning is performed via video-guided latent collocation with receding-horizon execution ($K=1$) under an action-conditioned world model. We solve the resulting constrained optimization using a primal--dual augmented Lagrangian method (ALM) with Adam updates, using $I_{\mathrm{ALM}}$ inner (primal) iterations per outer (dual) step and $O_{\mathrm{ALM}}$ outer iterations in total. Hyperparameters are selected using a small held-out validation set of 20 trajectories and are fixed across all experiments. For Push-T at horizon $T=25$, we tune the penalty growth factor $\gamma \in \{1.2, 1.3, 1.5, 1.7, 1.9, 2.2, 2.5, 2.9\}$, video alignment weight $\lambda_v \in \{0.01, 0.1, 1.0, 10.0, 20.0, 50.0\}$, action regularization weight $\lambda_r \in \{0.01, 0.05, 0.1, 0.5\}$, goal alignment weight $\lambda_g \in \{0.1, 1.0, 5.0, 10.0, 20.0, 50.0\}$, $I_{\mathrm{ALM}} \in \{10, 25, 50, 100, 200\}$, $O_{\mathrm{ALM}} \in \{5, 10, 25, 50\}$, and learning rate $\eta \in \{0.01, 0.05, 0.1, 0.5, 1.0\}$. The final Push-T configuration uses $\lambda_r=0.05$, $\lambda_v=1.0$, $\lambda_g=10.0$, $I_{\mathrm{ALM}}=25$, $O_{\mathrm{ALM}}=25$, $\rho_0=1.0$, and $\gamma=1.9$, with optional local refinement using 500 samples with variance $0.3$. For Push-T horizons $T \in \{50,80\}$, we set $\lambda_r=0.1$. For Wall, we use the same configuration as Push-T, except setting $\gamma=1.5$ for $T=25$; $T=50$ matches the Push-T configuration.

\end{document}